---

# Gender classification by means of online uppercase handwriting:
# A text-dependent allographic approach

Enric Sesa-Nogueras[*], Marcos Faundez-Zanuy, Josep Roure-Alcobé
Escola Superior Politècnica
TecnoCampus Mataró-Maresme

## Abstract

This paper presents a gender classification schema based on online handwriting.
Using samples acquired with a digital tablet that captures the dynamics of the writing, it classifies the writer as a male or a female. The method proposed is allographic, regarding strokes as the structural units of handwriting. Strokes performed while the writing device is not exerting any pressure on the writing surface, pen-up (in-air) strokes, are also taken into account. The method is also text-dependent meaning that training and testing is done with exactly the same text. Text-dependency allows classification be performed with very small amounts of text. Experimentation, performed with samples from the BiosecurID database, yields results that fall in the range of the classification averages expected from human judges. With only four repetitions of a single uppercase word, the average rate of well classified writers is 68%; with sixteen words, the rate rises to an average 72.6%. Statistical analysis reveals that the aforementioned rates are highly significant. In order to explore the classification potential of the pen-up strokes, these are also considered. Although in this case results are not conclusive, an outstanding average of 74% of well classified writers is obtained when information from pen-up strokes is combined with information from pen-down ones.

**Keywords: biometric recognition, gender recognition, handwriting, stroke, pen-up stroke, pen-down stroke**

## 1. Introduction

Most applications of biometry fall in the field of security. But biometric techniques and methods can be successfully applied to other fields ([1], [2]). Gender classification is one of these applications. Gender classification refers to the recognition of an individual's gender. The term gender means the physical and/or social condition of being male or female. Gender may be regarded as a demographic category, with age, handedness or ethnicity being instances of other such categories.

---
[*] Corresponding author. Tel.: +34 93 169 65 00; Fax: +34 93 169 65 05
Postal address: Escola Universitària Politècnica de Mataró. TecnoCampus Mataró-Maresme (Edifici universitari). Av. Ernest Lluch, 32 08302 Mataró (Spain).
e-mail addresses: sesa@tecnocampus.cat (E. Sesa-Nogueras); faundez@tecnocampus.cat (M. Faundez-Zanuy); roure@tecnocampus.cat (J. Roure-Alcobé)

From a biometric perspective, gender classification can be mainly performed based on physiological traits (e.g. face) or on behavioural traits (e.g. handwriting). Also "soft" traits (e.g. clothing) can give clues to the gender of a person.

Most of the published studies in the literature are focused on face ([3], [4] ,[5]) or speech ([6], [7]). Other traits, like gait ([8]), have drawn some attention too. Combinations of two or more traits, for instance face and fingerprint ([9]) or face and gait ([10]), have also been considered.

When performed by human beings, the accuracy of gender classification depends highly on the trait under consideration. In some cases an almost perfect rate is achieved. For instance, according to [11], human observers can classify photographs of non familiar faces with 96% accuracy. But in other cases, classification accuracy tends to be far from 100%. In the particular case of handwriting, the upper limit seems to be somewhere around 70%. More on how well human beings judge gender from handwriting will be said in section 2.

In scientific literature, not all biometric traits have been given the same importance when endeavouring to solve the gender-recognition problem. This translates into the fact that little is known on the potential of some traits for automatic recognition. One among these traits is handwriting. Thus, while established evidence exists that automatic gender-recognition can be performed using face or speech ([3], [12]), much less evidence exists pointing towards the existence of such a gender-recognition potential in handwriting. To some, the latter affirmation may come as a surprise since in some contexts it is pretty strongly believed that handwriting conveys a considerable amount of gender-related information. Some papers, a selection of them commented upon in the next section, seem to back up this belief when the classification is performed by human judges. But when it comes to automatic classification, a review of the existing literature quickly reveals a notable scarcity of papers dealing with this topic. Thus little scientific evidence backs up the aforementioned belief in the particular case of automatic computer-based classification. Nevertheless some papers exist, mainly in the offline field, and they will be reviewed in section 3.

Handwriting-based automatic gender classification has several fields of application: human-computer interaction, forensics, psychology and security, among others. In human-computer interaction it may help systems to guess their users' genders in order to render a more personalized interaction offering, for instance, gender-targeted advertisements. In forensics, it can help investigators to narrow the range of candidates to be investigated. In psychology, it may help the ongoing research on the correlation of the masculine/feminine writing  style and the ratio of second to fourth digit length (2D:4D ratio) which is seemingly influenced by sex hormones, mainly testosterone ([13]). Some researchers also link these results to the economic behaviour of people ([14]). In general, gender classification may enhance the performance of handwriting-based diagnose-aid tools since some conditions revealed by handwriting happen to have gender-dependent effects. Finally, biometric security applications may also benefit from automatic handwriting verification and gender classification since handwriting can be combined with other biometric modalities in order to achieve enhanced performances.

---

The work presented in this paper revolves around a gender recognition system that classifies handwriting according to the sex of the writer. Experimentation is carried out thanks to the handwriting samples in the BioscurID database, donated by 400 anonymous donors. The size of this database allows a reliable statistical analysis of the significance of the results. In order to support the claim that the proposed systems yields classification rates that are as high as those that could be expected from human judges, a whole section of the paper is devoted to reviewing the existing literature dealing with non-automatic gender classification.

One of the novelties of the proposed schema lies in the allographic nature of the system. It does not focus on global statistical features extracted from whole blocs of the text, as structural approaches do ([15]) but, instead, it regards strokes as the structural units of handwriting. A stroke may be a whole character, a fragment of single character or a fragment of handwriting comprising parts of more than one character. During the training stage, two sets of prototypical strokes are built: one from females' samples and one from males'. During the testing (classification) stage, a writer's handwriting is attributed the gender of the set that best matches his/her strokes. Another novel aspect of the proposed schema is its text-dependent nature: the text used in the testing phase has to be the same that was used in the training stage. In handwriting-based biometric systems, text-independency tends to require considerable amounts of text ([16]) while text-dependent systems can yield reasonable results with very small amounts of handwriting, even just a word or a few words ([17]). While text-independency may be of interest in forensic applications, text-dependency requiring but a smaller amount of text may be of interest in applications that require low-intrusiveness means of guessing its user's gender. Furthermore, the text-dependent nature of the proposed system makes it suitable to experiment with databases that contain but very short amounts of handwriting from each writer or the very same text for all of them.

Another aspect worth mentioning is that experimentation is carried out only with uppercase script. It is generally assumed that uppercase characters are poorer in writer-specific information than cursive lowercase ones ([18]). This makes the uppercase-based classification a more challenging problem. To the best knowledge of the authors of this paper, no attempts have been previously made to classify gender from uppercase samples.

Being based on data that contains time-dependent information, the approach of the proposed system is online. Although much of the existing literature deals with off-line approaches, mostly based on scanned images, the advent of devices equipped with touch screens that can acquire online handwriting with high accuracy allows us to foresee a growing interest in the online approaches.

And, finally, taking advantage of the online nature of the system the issue of the classification power of the in-air information (pen-up strokes) is put forward and, to the authors' best knowledge, investigated for the first time in the literature.

The outline of this paper is as follows: the next section highlights some relevant papers and results regarding non-automatic (i.e. carried out by human judges) gender recognition from handwriting. An important conclusion that can be drawn from them is stressed: there seems to be an upper limit for the accuracy of such recognition. Section 3 briefly reviews and summarizes the few existing papers that address the issue of automatic gender recognition from handwriting. Section 4 describes the classification method proposed, emphasizing its allographic nature. In section 5 the experimental setup and the obtained results are presented and its significance is discussed. Finally, section 6 concludes the paper, summarizing the main results achieved, also pointing out some inconclusive results regarding the potential of pen-up strokes that could lead to further research efforts.

## 2. Highlights on non-automatic gender recognition from handwriting

Psychologists and graphologists have been considering the question of whether gender can be determined accurately from handwriting since, at least, the first years of the previous century ([19]). The following paragraphs give an account of some relevant results obtained since the early twentieth century. These results seem to provide evidence that:

(a) Handwriting is far from being gender-neutral, i.e., males and females tend to write differently but

(b) There may be an upper limit regarding the accuracy with which human judges can establish whether a sample of handwriting has been produced by a male or a female.

In 1910 Downey ([20]) performed the following experiment: 200 envelopes, half of them written by women, were sent to 13 individuals, none of them having a background in graphology. They had to record their judgement as to the sex of the writer who had addressed each envelope. The percentages of well classified envelopes ranged from 60% to 77.5% (mean being 67.3). With this experiment, Downey was trying to replicate a previous experiment by M. Alfred Binet, a French psychologist who, in 1906, had claimed that sex could be determined accurately in 75% of the cases ([21]).

Later, in 1926, Newhall undertook a similar experiment also using postal addresses from 200 envelopes, half of which had been written by women ([22]). The postal addresses were given to 92 volunteers who had to estimate the gender of the writer. As in Downey's experiment, all the judging subjects were unskilled with respect to handwriting analysis. The average accuracy for the 92 participants was 57.34% which was deemed significantly above chance.

In 1931, Young carried out yet another similar experiment ([23]): 50 subjects, claiming no skill in graphology, were given 50 samples of handwriting, half of them having been produced by female students and the rest by male students. Differently from the aforementioned experiments of Binet, Downey or Newhall, Young's judging subjects were aware that exactly half the samples had been produced by males and the rest by females. Correct judgement ranged from 48% to 72%, being 61% the mean performance.

Other similar experiments were conducted during the twenties and the thirties, all yielding average accuracies of about two thirds or slightly above. The interested reader may see, for instance, [24] (average success 68.4%), [25] (average success 68%) and [26] (average success 63%).

According to the reported results of those earlier experiments, there seems to be a "*slightly-above-two-thirds barrier*" limiting the accuracy human judges may attain when performing the gender-classification task. In 1961, Fluckiger et al. in a review of the experimental research in graphology conducted in the period 1930-1960 ([27]) stated that 71% is "*the usual degree of success in such studies*" referring to the determination from handwriting of the psychological masculinity-femininity dimension. In the same paper, it is also stated that when it comes to physical rather than psychological gender, the percentage of correct guesses achieved in the surveyed works varies from 60 to 70%.

In 1971, Hodgins ([28]) also expresses similar conclusions about the average accuracy although in his experiment the inter-judge variability is somewhat higher (accuracy ranging from 40 to 80%). In 1996 Hecker ([29]) in another survey of the studies conducted from 1906 to 1991 concludes that the mean accuracy reached in those studies was of 71.7%.

All the aforementioned references deal with Western script performed in Western countries. Very little is known on the impact of the culture in the accuracy of gender judgement since no experimentation involving non-western scripts has been reported. An exception to this is [30] where the results of an experiment involving 30 writers and 25 judges are described. In this experiment, no statistically significant difference was found between English and Urdu, both achieving similar levels of accuracy (about 68%).

It is not in the scope of this paper to analyze the reasons that may explain the differences in handwriting that male and female writers tend to exhibit. Nevertheless, it is interesting to offer a brief account of the *nature and nurture* explanatory axis.

When the biological dimension (nature) is considered, differences in muscular strength (higher in men) and in the physiology of the hand (smaller and with more fatty tissue in women), may have some relevance. For instance graphologists tend to consider that men exert higher pressure than women when handwriting ([19]). Some authors point out that fine motor coordination is better in females [31]. Also prenatal sex hormones may have an impact on handwriting style, as the study of Beech and Mackintosh ([13]) suggests.

Regarding the sociological axis (nurture), there may be cultural stereotypes, perpetuated through the schooling systems, that determine attitudes that end up having expression in the handwriting, with the resulting differences persisting throughout life (e.g. girls should be neat while boys should be firm, strong and write faster and scruffier [31]).

Finally, it must be said that although it is generally accepted that a female's handwriting is neater and rounder than a male's (see left side of figure 1) whereas a handwriting produced by a male tends to be scruffier, more angular and executed exerting more pressure on the writing surface (see right side of figure 1), different authors attribute different, sometimes contradictory, qualities to the handwriting of each gender. Thus, while some may claim that females' handwriting tends to be bigger others believe that it is smaller than males' [19]. Hecker analyzed several features related to qualitative aspects of the handwriting (size, slant and shape) and concluded that none can be clearly

___________________________________________________________________

attributed to one of the genders. Even if males and females may have different mean values for those features, variability is too large to allow a sound distinction [29].

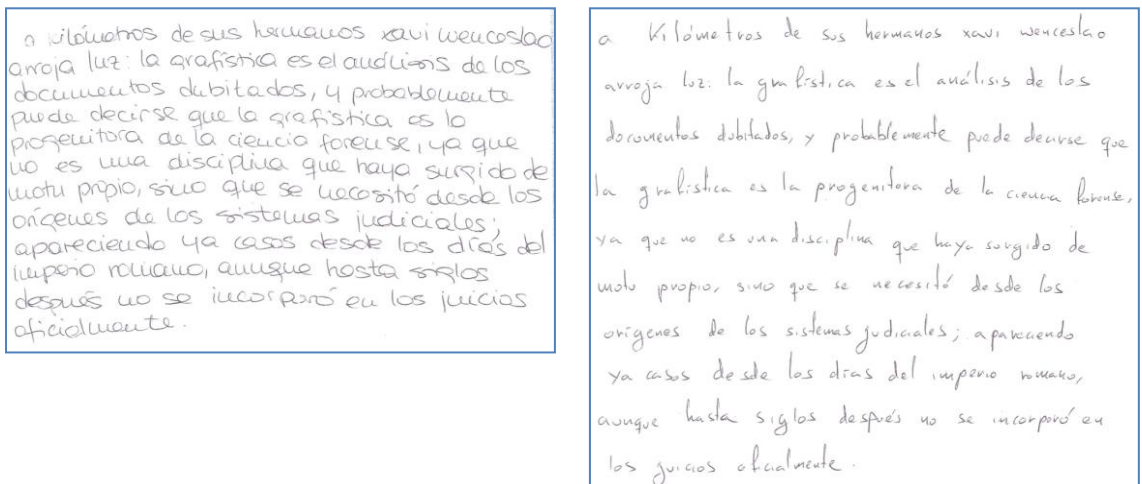

**Fig. 1:** 'Feminine' cursive handwriting produced by a woman (left) and 'Masculine' cursive handwriting produced by a man (right)

## 3. Related work: automatic gender recognition from handwriting

Contrary to other biometric modalities (e.g. face or voice), the matter of attempting to devise systems that automatically perform gender classification from handwriting has received but scant attention: the number of available references is really scarce, involving very few researchers. Nevertheless this situation may be changing and the near future may witness an increased interest in this field as the success of the gender prediction contest reported in [32], which attracted 194 participants, suggests.

In 1996, Hecker presented what may be one of the first attempts to use a computer assisted method to classify gender from handwriting. In [29], Hecker reports the results obtained with a semi-automatic classification approach. From digitized images from the FISH database ([33]), two sets of features were considered: the first consisted of automatically extracted ones (pixel autocorrelation, white pixel statistics, and black pixel statistics) whereas the second consisted of features measured semi-automatically with the help of a human operator (slant, shape of loop, ...) Experimenting with 96 males' and 96 females' handwritings the overall correct classification rate was 71.5%, being the best rate 72.4%.

Tomai et al., in [34], use a K-nn (K=6) classifier to analyze the discriminatory power of isolated characters with respect to several demographic characteristics. Actually the following groups are considered: gender, handedness, age (under 24 / above 45) and origin (US-born / non-US-born). The features considered are the *micro features* described in [35] and the samples used for training are taken from the well-known CEDAR letter database, also described in [35]. When it comes to gender classification,

___________________________________________________________________

the results are quite surprising since some individual characters show an unexpectedly high performance: letter *'b'* attains a 70% performance, digit *'3'* a 67%, letters *'m'* and *'y'* a 66% and letters *'Y'* and *'a'* a 65%. When classification is performed using all the available characters (10 digits, 26 lowercase and 26 uppercase letters, one sample per writer) the performance does not exceed 75%.

In [36], Bandi and Srihari tackle the gender-classification problem by means of the combination of up to 10 neural networks (NNs) using bagging and boosting techniques. In their experiments the NNs are trained with samples coming from 800 individuals and the testing phase is carried out with samples coming from 400 different individuals. All the samples come from the CEDAR letter database. The features used are 11 of the 12 *macro features* described in [35]. With a single NN the best accuracy attained is slightly over 73% while an accuracy of 77.5% (the best reported in that paper) is reached when applying boosting techniques to 10 NNs.

In [37], Liwicki et al. propose two different gender-classification systems: one based in support vector machines (SVMs) and the other based on Gaussian mixture models (GMMs). In both cases the overall approach is on-line since the systems are devised to be applied to data acquired by means of a digital whiteboard system ([38]). A combination of on-line features and off-line features is considered, the latter computed from a two-dimensional matrix representation of the on-line data. Experimentation is performed on the IAM-onDB database ([39]), with a training set consisting of samples from 100 individuals (50 women, 50 men) and a testing set of 50 individuals (25 women, 25 men). Accuracy ranges from 56.60%, when using a SVM with a sigmoid kernel, to 67.06%, when using a GMM.

The work presented in [37] is further elaborated in [40], by three of the four authors of the former paper. In the latter paper, the accuracies attained by a GMM-based system when considering the on-line features, the off-line ones and their combination are reported separately. Actually, it is reported that the on-line features achieve an accuracy of 64.25%, the off-line ones achieve a 55.39% and their combination raises the accuracy up to a 67.57%. Experimentation is also performed on the IAM-onDB database with a training set of 80 individuals and a testing set of 50, both gender-balanced.

In [32] Hassaïne et al. give details of a competition on gender prediction from offline handwriting hosted on Kaggle. All participants trained and tested their systems using data from a subset of the QUWI database ([41]) that contains handwriting samples in Arabic and English. 194 participants entered the competition with systems implementing a variety of classification techniques applied to a set of more than 30 features extracted from the scanned images of the samples [42]. An accuracy of around 70% is reported.

Al Maadeed and Hassaine successfully tackle the gender classification problem in the offline field (along with age and nationality classification) [43]. They propose and describe a set of novel features that are combined using several classifiers. Text-dependent and text-independent experiments on the QUWI database ([41]) are performed using Arabic and English samples. In Arabic, the best reported accuracy is 71.6% (text-independent, classification based on kernel discriminant analysis) while in English it is 74.7% (text-independent, classification based on random forests). Best accuracies in the case of text-dependent experiments are only slightly lower: 71.1% for both Arabic in English.

**Table 1** Concise summary of automatic gender recognition references

| Author(s) and reference(s) | Year | Approach | Classifier (s) | database | Size of training and testing sets (# of individuals) | Best reported accuracy |
|---|---|---|---|---|---|---|
| Hecker [29] | 1996 | Offline semi automatic | Not specified | FISH [33] | Testing: 192 | 72.4% |
| Tomai et al. [34] | 2004 | Offline | K-nn | The CEDAR letter database [35] | Training: 600 Testing: 300 | 70% (isolated letter '*b*') |
| Bandi & Srihari [36] | 2005 | Offline | Up to 10 combined neural networks | The CEDAR letter database [35] | Training: 800 Testing: 400 | 77.5% |
| Liwicki et al. [37] | 2007 | On-line (including off-line features) | Support Vector Machines Gaussian Mixture Models | IAM-OnDB database [39] | Training: 100 Testing: 50 | 67.06% |
| Liwicki et al. [40] | 2011 | On-line Off-line | Gaussian Mixture models | IAM-OnDB database [39] | Training: 80 Testing: 50 | 55.39% (offline) 64.25% (online) 67.57% (combination) |
| Competition on Gender Prediction from Handwriting [32] | 2013 | Off-line | Multiple | QUWI [41] | Training: 282 Testing: 193 | 70% |
| Al Maadeed & Hassaine [43] | 2014 | Off-line | Random Forests and Kernel Discriminant Analysis | QUWI [41] | Training: 712 Testing: 305 | 74.7% |

_________________________________________________________________

## 4. Classification method

### 4.1. *Approach*

The gender classification method presented in this paper follows a stroke-based approach. It relies on the particular election of strokes each particular writer makes and, more precisely, in the particular election each gender may make. It is in that sense that our method can be regarded as allographic in nature since it does not take into account the global characteristics of the text but focuses on strokes which may correspond to whole characters, character-fragments or shapes involving portions of several adjacent characters. Allographic systems have been very successfully used for writer identification and verification, both in the offline ([44], [45]) and in the online ([46], [17], [47], [48]) fields.

In our approach, sequences of text, which from now on will be referred to as *words*, are regarded as two separate subsequences: one of on-surface trajectories, or pen-down strokes, and another one consisting of in-air trajectories or pen-up strokes. A stroke is the trajectory and all the related data recorded by the acquisition device during the time that spans between a pen-down and a pen-up movement (or vice versa). Pen-down strokes correspond to the visible parts of the handwriting, while pen-up strokes are the zero-pressure in-air trajectories performed while transitioning from one pen-down stroke to the next. **Figure 2** shows pen-up and pen-down strokes from the execution of the Spanish word INEXPUGNABLE

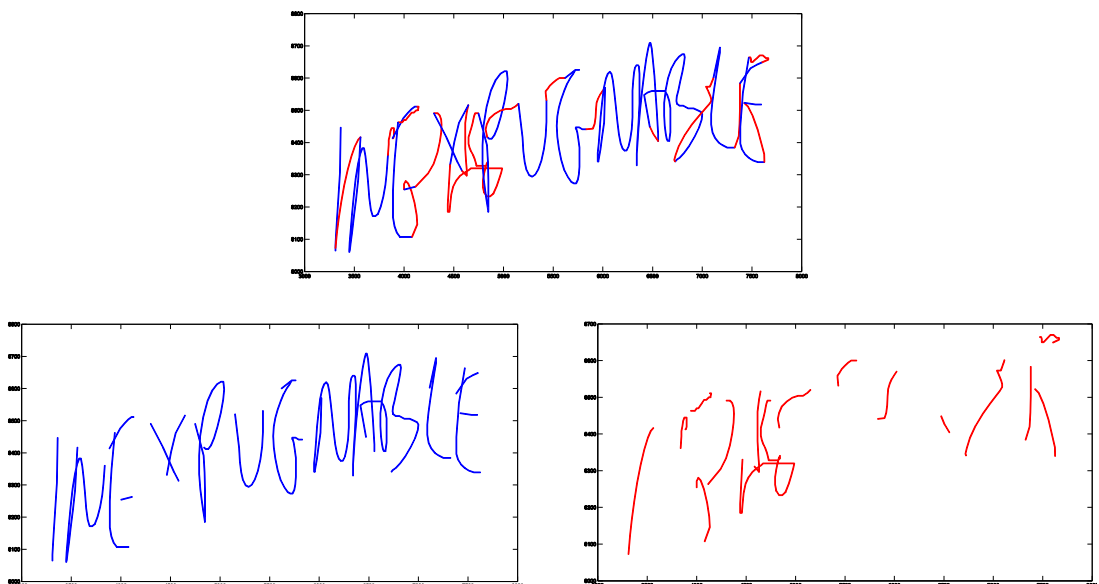

**Fig. 2**: Execution of the word INEXPUGNABLE as captured by the acquisition device. The image at the top shows both pen-up and pen-down strokes. The images at the bottom show pen-down (left) and pen-up (right) strokes separately.

_______________________________________________________________

### 4.2. Segmentation and pre-processing of strokes

The system implementing the classification method relies on online data that adheres to the SVC format [49]. In its raw format, each execution of a word is given as seven time-sequences (features): $x(t)$, the $x$ coordinate; $y(t)$, the y coordinate; $ts(t)$, a time stamp value; $bs(t)$, the button status value (0 for pen-up, 1 for pen-down); $az(t)$, the azimuth; $al(t)$, the altitude and $pr(t)$ the pressure. Thus, the execution of a word can be formally described as a matrix $[x(t), y(t), ts(t), bs(t), az(t), al(t), pr(t)]$ with $t \in [1, N]$ where N is the length (number of sampling units) of the execution. Segmentation into strokes is straightforwardly achieved thanks to the $bs(t)$ feature. A pen-down stroke starts at a point where $bs(t)$ changes from 0 to 1 and ends at a point where $bs(t)$ changes from 1 to 0. In pen-up strokes the $pr(t)$ feature can be discarded since it is always zero-valued.
In the experiment reported in this paper, the writing angles (azimuth and altitude) have been discarded since, in unpublished preliminary experiments performed by the authors, they appeared to be irrelevant.

The stroke pre-processing applied is essentially the same one described in [46]:

1. Non-selected features are removed
2. Each stroke is resampled to a fixed number of points since this is a requirement of the self-organizing maps (SOMs) used for gender modelling.
3. Selected features are normalized to mean 0 and standard deviation 1

Differently from [46], features are not given different weights since our preliminary unpublished experiments suggest they all have similar relevance.

This segmentation and pre-processing procedure is applied to all samples regardless of their posterior use (training or testing).

### 4.3. Training phase: gender modelling

The cornerstone of the classification method is a pair of codebooks of strokes, one for male writers, the *males-codebook*, and a different one for females, the *females-codebook*. A codebook is a set of classes, each one representing a whole set of similar strokes. During the training phase, the two codebooks are built in an unsupervised manner by means of SOMs ([50]). Samples come from male writers for the males-codebook and from female writers for the females-codebook. Once the training of the SOMs is completed, each one of their cells represents the prototype of a class of strokes. As a whole, a codebook from a particular gender can be regarded as the collection of strokes typically used by individuals of that gender. In other words, each gender's handwriting is modelled by a codebook of its prototypical strokes.

One of the aims of our work being to ascertain the gender-classification potential not only of visible trajectories, but also of the invisible in-the-air trajectories, a pair of codebooks is also built from pen-up strokes. Thus, two pairs of codebooks will be considered, one pair from pen-down strokes (*pen-down males-codebook* and *pen-down females-codebook*) and another pair from pen-ups (*pen-up males-codebook* and *pen-up*

_______________________________________________

*females-codebook*). **Figure 3** Provides a graphical depiction of the gender modelling process. **Figure 4** shows a portion of the pen-down codebooks obtained from samples of the word DELEZNABLE.

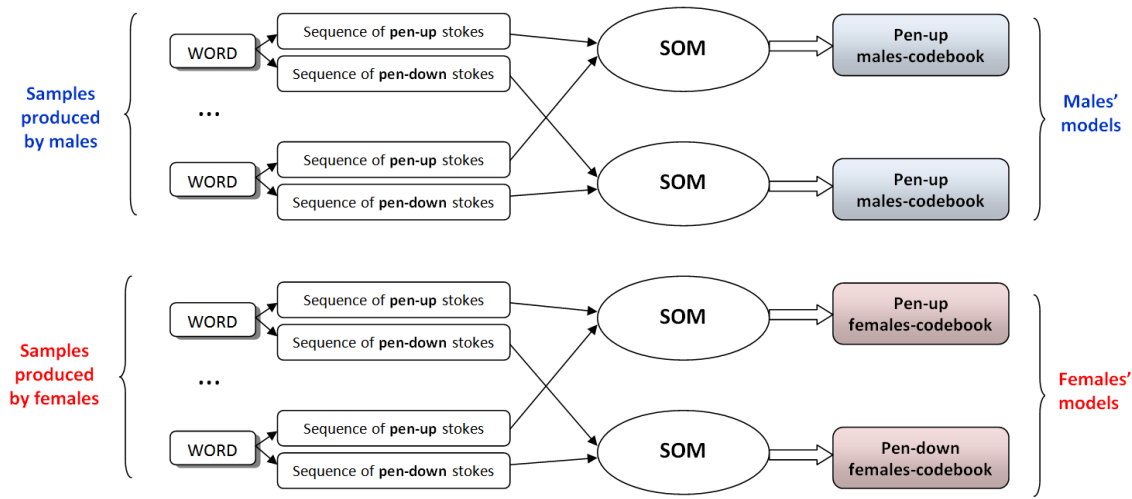

**Fig. 3:** Schematic overview of the construction of the models (training phase).

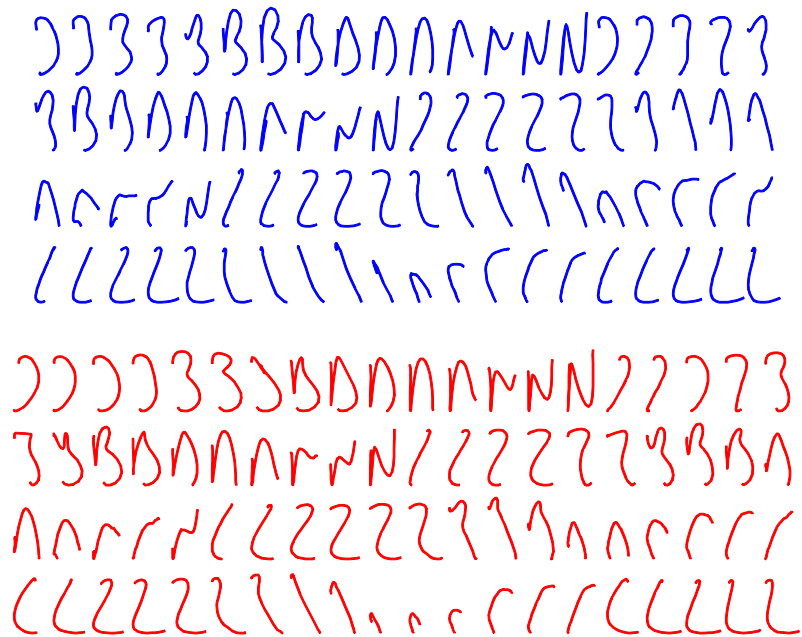

**Fig. 4:** Prototypical pen-down strokes obtained from samples of the word DELEZABLE. The four upper rows come from a *male-codebook* while the four bottom rows come from a *female-codebook*.

Training settings for the SOMs are exactly the same that were used in [46], namely: 150 units in a sheet-shaped two-dimensional lattice with a hexagonal neighbourhood topology. The final number of units may vary slightly because it is automatically optimized to improve the accommodation of the data presented to the SOM. The training is performed during 240 epochs: 40 rough-training epochs (high learning rate and high neighbourhood radius) and 200 fine-training epochs (lower learning rate and lower neighbourhood radius). The actual construction of the SOMs is left to the SOM Toolbox ([51], [52]), a specialized Matlab package.

Regarding time complexity, the training of a SOM takes a time that is linear in the number of data samples, linear in the number of training epochs and quadratic in the number of units. The number of units and the number of epochs being fixed, the codebooks can be built in a time linear in the number of strokes fed to the training algorithm. Using a conventional desktop computer equipped with a dual-core 2.7 Ghz processor and 8 Gb of RAM memory, the time taken to build a codebook from the strokes in 400 realizations of a word was about a second

### 4.4. Testing phase: gender classification

In order to determine the gender of an unknown writer a sample of his/her handwriting is taken. This being a text-dependent method, it is necessary that the sample contains the same text used in the previous phase (training). The sample provided is encoded (quantized) twice, first with the males' codebook and then with the females' one. For each codebook, the distortions caused by the quantization process are computed, on a per-stroke basis, and added together. Finally, the writer is attributed the gender of the codebook that produces the smallest distortion.

Classification can be performed threefold: (a) only considering the models obtained from pen-down strokes, (b) only considering the models obtained from the pen-up stokes or (c) considering both types of models by combining the measures (distortions) obtained from them. **Figure 5** is a graphical depiction of the pen-down-based classification mechanism.

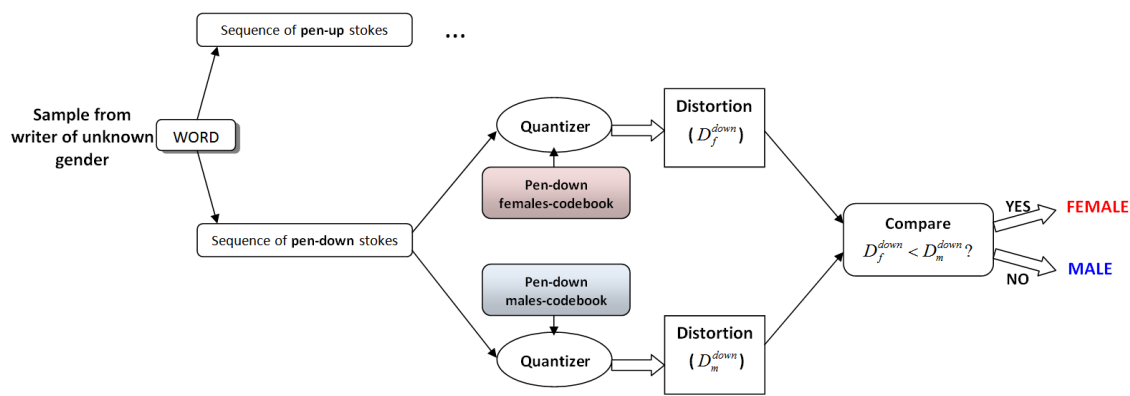

**Fig. 5**: Schematic overview of the classification process based on the codebooks (models) built from pen-down strokes. The same strategy is applied to pen-up strokes.

The computation of the distortion produced by a codebook when encoding a single word is linear in the number of strokes in that word. For practical purposes this time can be considered constant and almost negligible (using the configuration mentioned above, the time to compute a single distortion is in the order of milliseconds).

_______________________________________________________________

## 5. Experimental results

### 5.1. Database and settings

All the experimentation reported in this paper has been carried out using data from the BiosecurID database ([53]). This database comprises 8 different biometric traits, including handwritten text. All data was collected during 4 different sessions over a time span of 4 months. The number of donors was 400, with a balanced gender distribution. Regarding the handwritten text, each donor was requested to write 16 different Spanish words in uppercase, each one in a single line, without corrections or crossings out. The acquisition was carried out with a WACOM INTUOS A4 USB pen tablet.

30 of the 400 donors have been screened out because their handwriting did present corrections, crossings out or more than one word in a single line. Table 2 shows the 16 words that donors were requested to write and their lengths (number of letters). Table 3 summarizes the most relevant statistics regarding the participating donors. The whole set of available donors (370) contains 199 (53.8%) males and 171 (46.2%) females.

**Table 2** Words in the BiosecurID database

| WORD | TEXT | LENGTH |
|------|------|--------|
| W1 | BIODEGRADABLE | 12 |
| W2 | DELEZNABLE | 10 |
| W3 | DESAPROVECHAMIENTO | 18 |
| W4 | DESBRIZNAR | 10 |
| W5 | DESLUMBRAMIENTO | 15 |
| W6 | DESPEDAZAMIENTO | 15 |
| W7 | DESPRENDER | 10 |
| W8 | ENGUALDRAPAR | 12 |
| W9 | EXPRESIVIDAD | 12 |
| W10 | IMPENETRABLE | 12 |
| W11 | INEXPUGNABLE | 12 |
| W12 | INFATIGABLE | 11 |
| W13 | INGOBERNABLE | 12 |
| W14 | MANSEDUMBRE | 11 |
| W15 | ZAFARRANCHO | 11 |
| W16 | ZARRAPASTROSA | 13 |

**Table 3** Relevant statistics regarding the donors in the BiosecurID database

| GENDER DISTRIBUTION | | AGE DISTRIBUTION | | | | HANDEDNESS | |
|------|------|------|------|------|------|------|------|
| MALE | FEMALE | FROM 18 TO 25 | FROM 25 TO 35 | FROM 35 TO 45 | FROM 45 ONWARDS | RIGHT | LEFT |
| 54% | 46% | 42% | 22% | 16% | 20% | 93% | 7% |

A single experiment has been devised: 100 individuals, 50 males and 50 females are randomly chosen to build the models (codebooks), leaving 149 males and 121 females for testing. In order to have an equal number of individuals of each gender, 28 of the 149 males are randomly discarded. Then, testing is performed with 121 males and 121 females.

---

For each one of the 16 words four different codebooks are built, two for males one from pen-down strokes and one from pen-up strokes, and two for females, one from pen-down and one from pen up strokes (See Figure 4 in section 4.3.2). The 100 individuals chosen for model construction are the same for the 16 words. All 4 available sessions are considered, thus each codebook is built from 400 writing performances of the same word.

During the testing phase, for each donor and codebook the 4 different distortion measures obtained, one per session, are combined into a single distortion measure. Several combination strategies have been considered: the sum, the average, the maximum and the minimum and none has produced significantly better classification rates. Results reported in this section have all been obtained using the sum.

Four different instances of this experiment have been run (trial$_1$ to trial$_4$). Each trial considers different subsets of donors to build the models and perform the testing.

### 5.2. Human classification

For the sake of comparison, several people were asked to classify printed images of the 16 words produced by the donors (Figure 6). Actually two experienced professional graphologists and three computer-science academics with no particular training or interest in graphology. The professional graphologists refused to participate in this particular informal experiment alleging that uppercase script may not convey enough gender information making the classification harder and more unreliable.

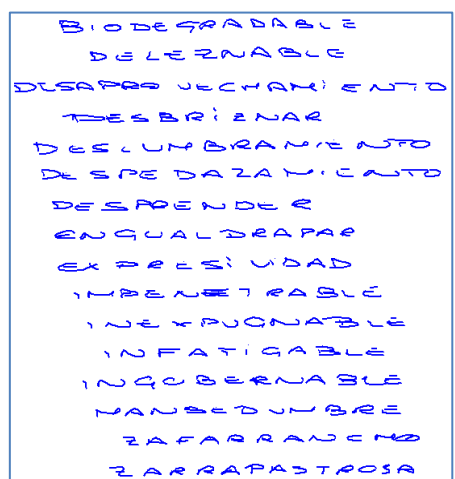

**Fig. 6**: **A sample of the uppercase words in the BiosecurID database produced by one of the anonymous donors.**

In order to explore the graphologists' claim that uppercase script is bound to produce worse classification results and also for the sake of comparison, cursive scripts, with the text shown in figures 1 and 2, and from the same donors, were also classified. This time the expert graphologists did not refuse to participate. Table 4 summarizes the results obtained from both types of handwriting.

**Table 4** Classification rates obtained from uppercase and cursive scripts by both graphologists and non-graphologists. All rates are statistically above chance.

| Classifier | Classification rate (Uppercase) | Classification rate (Cursive lowercase) |
|---|---|---|
| CS Academic 1 | 66.40% | 68.00% |
| CS Academic 2 | 60.00% | 64.80% |
| CS Academic 3 | 68.80% | 73.60% |
| Average (academics) | 65.07% | 68.80% |
| Graphologist 1 | | 68.80% |
| Graphologist 2 | | 68.80% |
| Average (graphologists) | | 68.80% |
| Average (all five) | | 68.80% |

Although no sound conclusions can be drawn from the results of this human classification, they are consistent with the claim that uppercase-based classification is a more challenging problem and is bound to produce lower rates. Also, it is consistent with the existence of an upper limit regarding how well humans can infer gender from handwriting samples, with this upper limit being slightly above two thirds, as has been previously commented on in section 2.

### 5.3. Results

#### 5.3.1. Classification rates achieved with each type of strokes.

The classification rates obtained in each trial are given in tables 5 and 6 for pen-down and pen-up strokes respectively.

**Table 5** Classification rates. Pen-down strokes only. All rates are statistically significant for $\alpha=99\%$

| PEN-DOWN STROKES | | | | | |
|---|---|---|---|---|---|
| WORD | TRIAL 1 | TRIAL 2 | TRIAL 3 | TRIAL 4 | AVG |
| W1 | 62,8% | 68,6% | 64,9% | 67,8% | 66,0% |
| W2 | 67,8% | 66,5% | 64,5% | 68,6% | 66,8% |
| W3 | 65,7% | 66,9% | 66,9% | 70,7% | 67,6% |
| W4 | 69,0% | 68,6% | 66,5% | 71,1% | 68,8% |
| W5 | 69,8% | 72,3% | 67,8% | 69,8% | 69,9% |
| W6 | 68,6% | 69,4% | 61,6% | 69,8% | 67,4% |
| W7 | 70,7% | 67,4% | 63,2% | 65,3% | 66,6% |
| W8 | 65,7% | 66,1% | 64,9% | 72,3% | 67,3% |
| W9 | 69,0% | 68,6% | 64,0% | 71,9% | 68,4% |
| W10 | 68,2% | 71,9% | 65,7% | 68,6% | 68,6% |
| W11 | 62,8% | 67,4% | 66,9% | 71,9% | 67,3% |
| W12 | 71,1% | 66,1% | 66,1% | 71,9% | 68,8% |
| W13 | 70,2% | 69,0% | 69,0% | 74,0% | 70,6% |
| W14 | 69,4% | 72,7% | 63,2% | 69,4% | 68,7% |
| W15 | 68,2% | 61,6% | 65,7% | 67,8% | 65,8% |
| W16 | 68,2% | 68,6% | 65,3% | 73,6% | 68,9% |
| AVG | 67,9% | 68,2% | 65,4% | 70,3% | **68,0%** |

_______________________________________________________________

**Table 6** Classification rates. Pen-up strokes only. Shadowed cells contain rates which are not statistically significant for α=99%

| | PEN-UP STROKES | | | | |
|---|---|---|---|---|---|
| **WORD** | **TRIAL 1** | **TRIAL 2** | **TRIAL 3** | **TRIAL 4** | **AVG** |
| W1 | 59,1% | 64,5% | 58,3% | 64,5% | 61,6% |
| W2 | 57,9% | 57,0% | 61,6% | 59,9% | 59,1% |
| W3 | 55,4% | 57,9% | 61,2% | 62,8% | 59,3% |
| W4 | 53,3% | 60,7% | 64,0% | 58,7% | 59,2% |
| W5 | 61,6% | 52,5% | 66,1% | 61,6% | 60,4% |
| W6 | 62,8% | 57,0% | 62,0% | 64,0% | 61,5% |
| W7 | 55,0% | 61,2% | 57,9% | 60,3% | 58,6% |
| W8 | 57,4% | 57,9% | 58,3% | 56,6% | 57,5% |
| W9 | 59,5% | 65,3% | 62,0% | 64,5% | 62,8% |
| W10 | 59,1% | 60,7% | 59,9% | 65,7% | 61,4% |
| W11 | 54,5% | 56,6% | 62,4% | 58,7% | 58,1% |
| W12 | 56,2% | 62,8% | 59,1% | 67,4% | 61,4% |
| W13 | 55,8% | 60,3% | 60,7% | 56,6% | 58,4% |
| W14 | 57,9% | 64,5% | 58,3% | 62,0% | 60,6% |
| W15 | 55,8% | 64,5% | 59,9% | 67,8% | 62,0% |
| W16 | 54,5% | 62,8% | 60,3% | 63,2% | 60,2% |
| **AVG** | 57,2% | 60,4% | 60,7% | 62,1% | **60,1%** |

First of all, these results have to be analyzed from the perspective of the statistical significance. What is the probability of obtaining these rates by pure chance (i.e. tossing a coin)? To answer this questions a straightforward binomial test ([54]) can be applied. If $r$ is the rate of well-classified writers, then null ($H_0$) and alternative ($H_1$) hypothesis can be stated as follows:

$H_0 : r = 50\%$
$H_1 : r > 50\%$

With the number of subjects considered in the experiments (242) and requiring a *p-value* $< 10^{-2}$ (significance level α=99%), the minimum rate that allows the rejection of the null hypothesis is $r_m = 57.8\%$. This means that the probability of observing a rate $r$ with $r \geq r_m$ would be less than 0.01 if the classification were purely random.

In table 6 (pen-up strokes), the rates below $r_m$ have been shadowed. In the case of pen-down strokes, all rates are statistically significant.

Notice that for pen-down strokes the overall average rate (68.0%) yields a *p-value* $\approx$ $8 \cdot 10^{-9}$ and that for pen-up strokes the overall average rate (60.1%) yields a *p-value* $\approx$ $8 \cdot 10^{-4}$ which are both well below the established *p-value* of $10^{-2}$.

It is worth noticing that the correlation between the lengths of the words and their classification rates is positive but very low (0.15 for pen-downs and 0.07 for pen-ups). Thus length seems to have a positive impact on classification rate but is far from

---

explaining the differences among words. In the case of pen-downs, the best performing word is *INGOBERNABLE* (70.6%, length 12) while the worst performing is *ZAFARRANCHO* (65.8%, length 11). Nevertheless the difference in classification accuracy between both words is not large: *ZAFARRANCHO* performs 6.8% worse than *INGORBERNABLE*.

For pen-ups, the best performing word is *EXPRESIVIDAD* (62.8%, length 12) while the worst performing is *ENGUALDRAPAR* (57.5%, length 12). Both words have the same length. When it comes to the correlation between the classification rates it is negative but negligible: -0.07. *ENGUALDRAPAR* has classification accuracy 8.44% lower than *EXPRESIVIDAD*.

### 5.3.2. Classification rates achieved when combining up and down strokes

In our past research ([46], [17], [55], [56]), it was shown that for identification purposes (identity recognition and identity verification) pen-up and pen-down strokes contain a certain amount of non-redundant information. This means that measures obtained from each type of stroke can be combined in order to obtain enhanced performances.
Table 7 summarizes the average classification rates obtained by the combination of the measures (distortions) obtained from the pen-down and pen-up models. The combination strategy was the non-weighted sum of the distortions. Other strategies were considered but none yielded significantly better rates than those shown in table 7.

**Table 7** **Classification rates for pen-down and pen-up strokes and their combination. All figures are averages over the four trials.**

| WORD | PEN-DOWN (AVG) | PEN-UP (AVG) | COMBINATION (AVG) |
|------|------|------|------|
| | PEN-DOWN + PEN-UP STROKES | | |
| W1 | 66,0% | 61,6% | 67,4% |
| W2 | 66,8% | 59,1% | 67,4% |
| W3 | 67,6% | 59,3% | 68,7% |
| W4 | 68,8% | 59,2% | 69,3% |
| W5 | 69,9% | 60,4% | 72,1% |
| W6 | 67,4% | 61,5% | 67,5% |
| W7 | 66,6% | 58,6% | 67,0% |
| W8 | 67,3% | 57,5% | 66,7% |
| W9 | 68,4% | 62,8% | 69,2% |
| W10 | 68,6% | 61,4% | 69,1% |
| W11 | 67,3% | 58,1% | 68,1% |
| W12 | 68,8% | 61,4% | 70,0% |
| W13 | 70,6% | 58,4% | 70,9% |
| W14 | 68,7% | 60,6% | 68,1% |
| W15 | 65,8% | 62,0% | 66,1% |
| W16 | 68,9% | 60,2% | 68,9% |
| AVG | 68,0% | 60,1% | **68,5%** |

_______________________________________________________________

The best performing word is *DELEZNABLE* (72.1%, length 10) while *ZAFARRANCHO* (66.1%, length 11) is the word that performs worst. The word *ZAFARRANCHO* has classification accuracy 8.3% lower than *DELEZNABLE*. This time the correlation between the classification rates and the length of the words is negative yet very small in absolute value: -0.16. Again, length does not seem to have any explanatory power regarding the classification performance.

According to the results shown in table 7, the combination of both types of strokes does not yield higher rates than those obtained from pen-down strokes only. Although it is out of the scope of this paper to provide a conclusive explanation of this fact, two conjectures can be made: it could be that, regarding gender, pen-up and pen-down strokes contain highly redundant information, or it could be that the combination strategy chosen is not appropriate to obtain enhanced rates. Further research will be focused on different and more sophisticated combination strategies in order to explore this issue. Nevertheless, it is important to point out that both types of trajectories seem suitable for classification purposes.

### 5.3.3. Classification rates combining the sixteen words

In our past research ([46], [17]), it has been shown that although a single on-line word may contain enough information to identify its author with notable accuracy (close to the accuracy achieved with signatures), the combination of several words yields even more accurate results. A similar behaviour can be expected regarding gender recognition.

Tables 8, 9 and 10, summarize the results obtained when the whole set of sixteen words in the BiosecurID database is considered and their sixteen individual measures (distortions) are combined into a single one. Again, the combination strategy is the straightforward non-weighted addition of the individual measures.

**Table 8** Classification rates for pen-down strokes with one (average) and all sixteen words in the BiosecurID database.

| PEN-DOWN STROKES | | | | | |
|---|---|---|---|---|---|
| | TRIAL 1 | TRIAL 2 | TRIAL 3 | TRIAL 4 | AVG |
| ONE WORD ONLY (AVERAGE) | 67,9% | 68,2% | 65,4% | 70,3% | **68,0%** |
| ALL SIXTEEN WORDS | 73,6% | 72,3% | 69,0% | 75,6% | **72,6%** |

**Table 9** Classification rates for pen-up strokes with one (average) and all sixteen words in the BiosecurID database.

| PEN-UP STROKES | | | | | |
|---|---|---|---|---|---|
| | TRIAL 1 | TRIAL 2 | TRIAL 3 | TRIAL 4 | AVG |
| ONE WORD ONLY (AVERAGE) | 57,2% | 60,4% | 60,7% | 62,1% | **60,1%** |
| ALL SIXTEEN WORDS | 58,7% | 65,7% | 61,2% | 69,4% | **63,7%** |

**Table 10** Classification rates for the combination of pen-down and pen-up strokes with one (average) and all sixteen words in the BiosecurID database.

| PEN-DOWN + PEN-UP STROKES | | | | | |
|---|---|---|---|---|---|
| | TRIAL 1 | TRIAL 2 | TRIAL 3 | TRIAL 4 | AVG |
| ONE WORD ONLY (AVERAGE) | 67,8% | 69,2% | 66,4% | 70,8% | **68,5%** |
| ALL SIXTEEN WORDS | 72,7% | 74,4% | 70,7% | 78,1% | **74,0%** |

In all cases, the greater amount of information made available to the classifier translates into a higher classification rate.

## 6. Conclusions

In this paper a novel method for performing gender recognition from online handwriting samples has been proposed. This novel method, based on an allographic approach and using uppercase text, achieves results that are as good as other results reported in the literature: using only pen-down strokes (x-y trajectories and pressure), an average classification rate of 68% is obtained with a single word (ranging from 65.8% to 70.6% depending on the particular word considered). This classification rate is slightly above the 64.25% and the 65.57% reported in [40]. However, caution should be exercised here since these figures were obtained from different databases and methodologies thus thwarting the possibility of straightforward direct comparisons.

A classification rate of 68%, achieved when testing with 242 individuals, yields a very low p-value ($\approx 8 \cdot 10^{-9}$) thus conferring a high degree of statistical significance upon the performance of the classifier. It is also worth noticing that all the 64 experiments performed with pen-down strokes yielded statistically significant rates. Moreover, the results achieved with a single word seem to be in accordance with the results that would be expected if classification were performed by human judges. In this sense, the classification method presented in this paper may be claimed to perform as accurately as an average human judge would (the reader may recall, from section 2, the "*slightly-above-two-thirds barrier*").

Regarding the classification power of pen-up strokes, the results are somehow less conclusive. When only this type of strokes is considered, the classification rates and their statistical significance drop considerably. With a single word, an average 60.1% of well classified writers is achieved. Although this average figure is statistically significant (*p-value* $\approx 8 \cdot 10^{-4}$), 15 out of the 64 experiments performed yielded non-significant classification rates with a p-value higher than the required $10^{-2}$ for α=99% (see shadowed cells in table 6). When not one but the sixteen words are considered, then classification rates are all significant but still below the rates achieved with pen-down strokes.

As for the source(s) of the differences between the two types of strokes, it is completely out of the scope of this paper to reach any sound conclusion. It may well be that pen-up strokes contain significantly less gender-related information than pen-down strokes,

---

although in the field of writer identification and writer verification they can perform almost equally well ([46], [17], [55]). The source of the difference could also lie with the classification method itself, although the very same method achieves notably good results with pen-down strokes.

When both pen-down and pen-up strokes are considered, combining their separate measures into a single measure, the classification rates do not improve significantly if only one word is considered: from an average 68.0% with pen-down strokes only to an average 68.5% with both types of strokes. When the sixteen words are all considered together, the improvement is significant but not outstanding: from an average 72.6% to an average 74%. This could be due to a considerable redundancy between both types of strokes, to the combination strategy chosen or to other causes. Again, it is out of the scope of this paper to cast any conclusive light on the reasons leading to the observed lack of improvement.

Another aspect that is worth noticing is the very limited amount of information required from each writer. With only four repetitions of a single uppercase word a "human-like" and state-of-the-art performance (68%) is already achieved. When more information is considered, i.e. sixteen words per writer, all the classification rates are higher than 70% with an outstanding 74% average. When compared to state-of-the-art results in the offline case, these figures are slightly above those obtained in [32] and very similar to those in [43]. Again, much care must be taken when extracting conclusions from the comparison of these figures since they were obtained from different databases and methodologies.

Summarizing, it can be concluded that the proposed classification method yields state-of-the-art, or higher, classification rates using a limited amount of information obtained from capital letters, which the expert graphologists consider more challenging than the lowercase letters. It adds evidence to the feasibility of automatic handwriting-based gender recognition, a topic on which little research has been published. It also contributes to the acknowledgement that handwriting contains gender-specific characteristics that can be traced back to the individual strokes that make up letters and words. Finally, for the first time in the literature the gender-specificity of the in-air trajectories that modern devices can acquire has been explored although, in this particular case, no conclusive evidence has been found.

**Acknowledgement**

We would like to acknowledge the graphologists Mari Luz Puente and Francesc Viñals for their support classifying data. This work has been supported by Ministerio de Economía y Competitividad TEC2012-38630-C04-03, and European COST action IC1206.

---

---